\pdfoutput=1

\documentclass[11pt]{article}

\usepackage[]{ACL2023}

\usepackage{times}
\usepackage{latexsym}

\usepackage[T1]{fontenc}

\usepackage[utf8]{inputenc}

\usepackage{microtype}

\usepackage{inconsolata}
\usepackage{amsmath}
\usepackage{breqn}
\usepackage{color,soul}
\usepackage{xspace}
\usepackage{multirow}
\usepackage{booktabs}
\usepackage{caption}
\usepackage{subcaption}
\usepackage{soul}

\newcommand{\knn}{\textit{KNN}\xspace}
\newcommand{\kf}{\textit{KNN-Former}\xspace}
\newcommand{\declarelogo}[0]{\includegraphics[height=.02\textwidth]{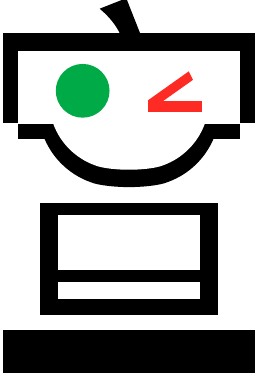}}

\title{Lightweight Spatial Modeling for Combinatorial Information Extraction From Documents}

\author{Yanfei Dong\(^{12}\), Lambert Deng\(^{1}\), Jiazheng Zhang\(^1\)\\ \textbf{Xiaodong Yu\(^1\), Ting Lin\(^1\), Francesco Gelli\(^1\), Soujanya Poria$^{\declarelogo}$, Wee Sun Lee\(^2\)}\\
\(^1\) PayPal  \(^2\) National University of Singapore\\
$^{\declarelogo}$ DeCLaRe Lab, Singapore University of Technology and Design\\
\fontsize{10}{10}\texttt{\{dyanfei, yuadeng, jiazzhang, xiaodyu, tinlin, fgelli\}@paypal.com}\\ 
\fontsize{10}{10}\texttt{sporia@sutd.edu.sg, leews@comp.nus.edu.sg}
}

\begin{document}
\maketitle
\begin{abstract}
Documents that consist of diverse templates and exhibit complex spatial structures pose a challenge for document entity classification. We propose \kf, which incorporates a new kind of spatial bias in attention calculation based on the K-nearest-neighbor (\knn) graph of document entities. 
We limit entities' attention only to their local radius defined by the \knn graph. We also use combinatorial matching to address the one-to-one mapping property that exists in many documents, where one field has only one corresponding entity. 
Moreover, our method is highly parameter-efficient compared to existing approaches in terms of the number of trainable parameters. Despite this, experiments across various datasets show our method outperforms baselines in most entity types. 
Many real-world documents exhibit combinatorial properties which can be leveraged as inductive biases to improve extraction accuracy, but existing datasets do not cover these documents.
To facilitate future research into these types of documents, we release a new ID document dataset that covers diverse templates and languages. We also release enhanced annotations for an existing dataset.\footnote{\url{https://github.com/miafei/knn-former}}

\end{abstract}

\section{Introduction}
Structured document information extraction (IE) attracts increasing research interest due to the surging demand for automatic document processing, with practical applications in receipt digitization, workflow automation, and identity verification etc.

Recent state-of-the-art methods for processing documents with complex layouts extensively exploit layout information, such as position, relative distance, and angle, with transformer-based models. 
Spatial modelling is a key contributing factor to the success of these methods (~\citealt{10.1145/3394486.3403172}, ~\citealt{appalaraju2021docformer}, ~\citealt{xu-etal-2021-layoutlmv2}, ~\citealt{hwang-etal-2021-spatial}). However, absolute coordinates, pair-wise relative Euclidean distance, and angle are insufficient to capture the spatial relationship in complex layouts. Two document entity pairs could carry different importance despite having the same position and distance, due to the presence or absence of other entities positioned between the pairs. We believe that spatial information can be better exploited for document entity classification. 

We propose \kf, a parameter-efficient transformer-based model that extracts information from structured documents with combinatorial properties. In addition to relative Euclidean distance and angle embeddings as inductive biases~\cite{hwang-etal-2021-spatial}, we introduce a new form of spatial inductive bias based on the K-Nearest Neighbour (\knn) graph which is constructed from the document entities and integrate it directly into the attention mechanism. Specifically, we first construct a \knn graph based on the relative Euclidean distance of document entities. Then we incorporate hop distance between entities, which is defined as the shortest path between two entities on the \knn graph, in training their pair-wise attention weight. For entity pairs with the same Euclidean distance but different hop distance, the difference in hop distance would still contribute to different attention weights. 
 We limit an entity's attention calculation only to its local radius of neighborhood defined by the \knn graph. This also strengthens the inductive bias as reflected by our experiment results.

Furthermore, many real-world document information extraction tasks come with combinatorial properties, such as one-to-one mapping between field categories and values. Such combinatorial properties can be leveraged as inductive biases to improve the extraction accuracy, but are under-explored because existing datasets do not cover such documents. 
Current methods that do not address the combinatorial constraints suffer suboptimal performance on these types of documents.
 
We further leverage this inductive bias by treating the entity classification task as a set prediction problem and using combinatorial matching to post-process model predictions\cite{kuhn1955hungarian, carion2020end,stewart2016end}.

In addition, \kf is parameter-efficient. Recent baseline models are initialized with parameters of pre-trained language models~\cite{10.1145/3394486.3403172, xu-etal-2021-layoutlmv2, hwang-etal-2021-spatial, hong2022bros}, making their model size larger or at least comparable to the language models. 

\kf does not utilize initialized parameters of existing language models, therefore free from the parameter size floor restriction. It is designed to be 100x smaller in trainable parameters compared to  prevailing baselines.
\kf's parameter efficiency makes it energy-efficient, contributes to faster training, fine-tuning and inference speed and 
makes mobile deployment feasible.

To encourage the progress of IE research in complex structured documents with combinatorial mapping properties, we release an ID document dataset (named POI). 
While the existing ID document dataset has only 10 templates \cite{DBLP:journals/corr/abs-2107-00396}, POI exhibits better template and lingual diversity. It also has a special mapping constraint where one field category has only one corresponding entity. 
In compliance with privacy regulations, the documents in the POI dataset are specimens and do not contain information about real persons. 

We conduct extensive experiments to evaluate the effectiveness of our proposed method. 
\kf outperforms baselines on most field categories across various datasets, despite having a significantly smaller model size. Extensive ablation studies show the importance of the \knn-based inductive bias and combinatorial matching.

To summarize, our contributions include (1) a highly parameter-efficient transformer-based model that (2) incorporates \knn-based graph information in local attention; (3) combinatorial matching to address the one-to-one mapping constraint; (4) a new ID document dataset with good template diversity, complex layout, and a combinatorial mapping constraint. 

\begin{figure*}[htbp]
\centering

\includegraphics[width=0.8\linewidth]{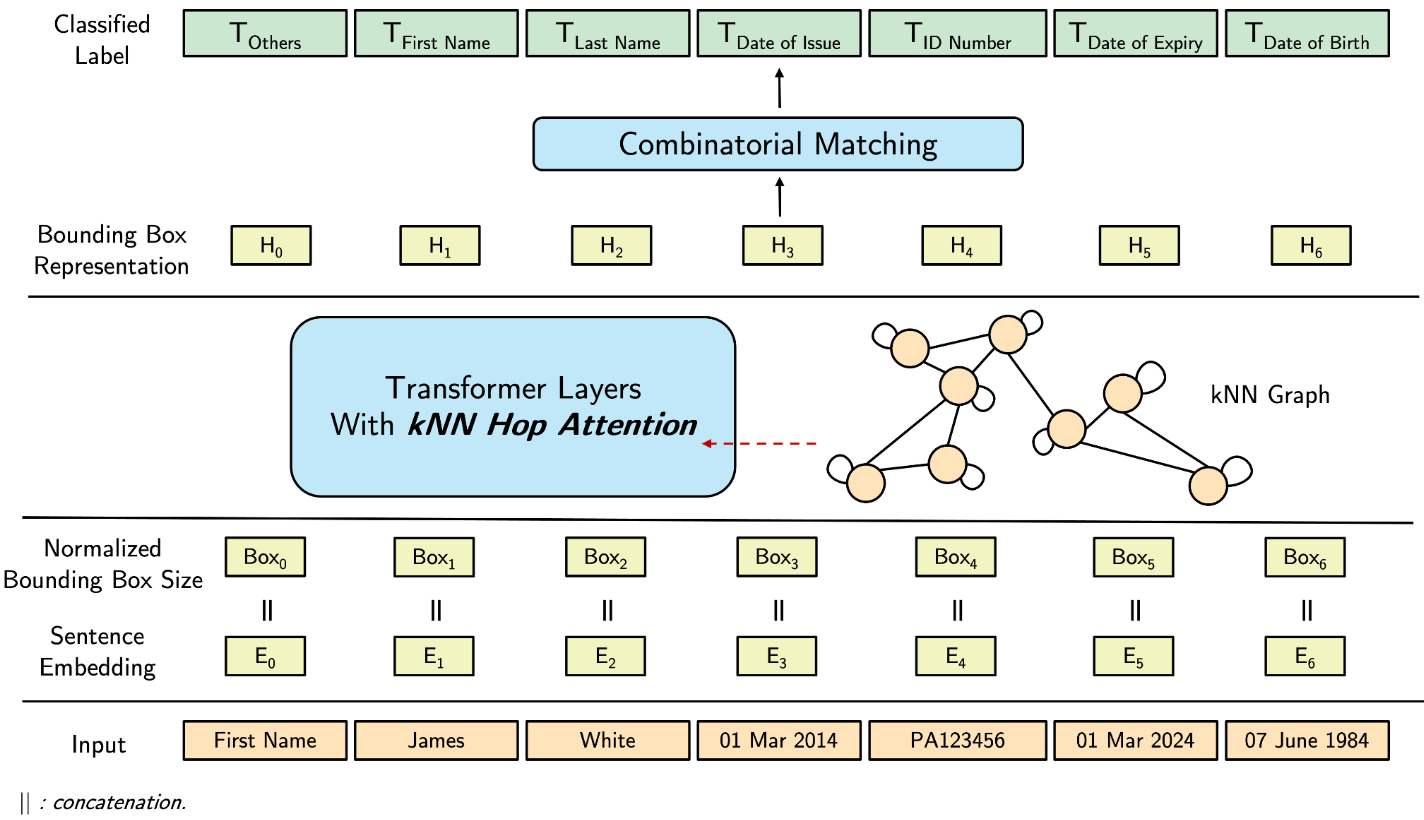}
\caption{An illustration of \kf. Bounding box texts are embedded using sentence transformer, which are concatenated with embeddings of bounding box size to form input embeddings. The concatenated embeddings are then passed to the transformer layers with \knn Hop Attention, which incorporates pair-wise relative hop distance between entities on \knn graph in attention calculation. 
The output entity representations of the transformer layers are passed to combinatorial matching for set prediction.}
\label{fig:archi}
\end{figure*}

\section{Related Work}

Researchers have tried multiple approaches for document information extraction~\cite{8892998,Mathew_2021_WACV, 10.1007/978-3-030-86549-8_36}.
However, these works do not have spatial cues, such as the position of the information in the original document.
To address this shortcoming, a number of works introduce the modality of layout information as additional input features. \citet{majumder-etal-2020-representation} adopts positional information as inputs to their method to extract information from receipt documents. LayoutLM~\cite{10.1145/3394486.3403172} adds 1-D and 2-D absolute position encodings to text embeddings before passing them to the transformer. \citet{DBLP:journals/corr/abs-2108-04539} proposes to train a language model from unlabeled documents with area masking, encoding relative positions of texts. StructuralLM~\cite{li2021structurallm} assigns the bounding box cell position as the position coordinates for each word contained in it. DocFormer~\cite{appalaraju2021docformer} encodes 2D spatial coordinates of bounding boxes for visual and language features. LayoutLMv2~\cite{xu-etal-2021-layoutlmv2} uses learnable pair-wise relative positional embeddings as attention bias.

A few works propose to use graphs to represent spatial entity relationships in documents. 
SPADE~\cite{hwang-etal-2021-spatial} uses a three steps graph decoder and formulates the information extraction task as a dependency parsing problem. 
FormNet~\cite{lee-etal-2022-formnet} constructs a k-nearest neighbor graph and applies a 12-layer graph convolutional network (GCN) to get the entity embeddings before feeding them into a transformer network. However, there are some limitations in using GCN to obtain embeddings. It is well established that the message passing-based GCN are limited in their expressive power~\cite{xu2018powerful, arvind2020weisfeiler, morris2019weisfeiler, chen2020can, loukas2019graph, dehmamy2019understanding}. In addition, FormNet does not use the hop distance between nodes, which could serve as a strong inductive bias to capture the spatial relationships between document entities. 

Datasets with positional information such as Funsd~\cite{8892998}, Cord \cite{park2019cord}, Sroie~\cite{huang2019icdar2019} are released to facilitate research in document understanding. However, they do not contain documents with combinatorial properties which are common in real-world applications.
MIDV\-500 \cite{DBLP:journals/corr/abs-1807-05786} and
MIDV\-2020 \cite{DBLP:journals/corr/abs-2107-00396} are two synthetic ID datasets with combinatorial properties, but are unsuitable for document information extraction tasks due to incomplete annotations. They also lack template diversity.

\section{Methodology}
In this section, we discuss the methodology for our model. We formulate the problem in Sec.\ref{sec:prob_formulation} and explain our overall model architecture and the details of each component in Sec.\ref{sec:model_archi}.

\subsection{Problem Formulation} 
\label{sec:prob_formulation}
Given a document $D$ which consists of multiple entities $\left\{e_i, \dots, e_j\right\}$, 
and the bounding box coordinates and texts$\left\{x_i, \dots, x_j\right\}$ 
detected by Optical Character Recognition (OCR) tool. 
We measure the relative distance and angle between two entities $e_i$ and $e_j$ 
as $\sigma_{\left(i, j\right)}$ based on the coordinates of bounding boxes.
Our task is to map each entity $e_i$ in document $D$ to its field category $y_i$, which is one of the predefined labels. For each field category $y_i$, there is only one corresponding entity $e_i$.

\subsection{Model Architecture}
\label{sec:model_archi}

We propose \kf, a transformer-based model for document entity classification. The architecture of \kf is shown in Fig. \ref{fig:archi}. \kf uses K-Nearest Neighbours Hop Attention, which incorporates a new inductive bias into attention computation. \kf also treats document entity classification as a set prediction problem and uses combinatorial assignment to address the one-to-one correspondence between entities and fields. \kf is highly parameter-efficient compared to baselines. Details of model size can be found in Tab~\ref{tab:main_table}.

\subsubsection{K-Nearest Neighbors Hop Attention} 

One key contribution of \kf is the proposed attention mechanism. Following~\cite{lee-etal-2022-formnet}, we first construct a \knn graph based on the Euclidean distance between each pair of entities. We represent entities as nodes and then connect edges between each entity and its \textit{K} nearest neighboring entities. We also add a self-loop to each entity to improve performance~\cite{kipf2016semi}.
While previous works focus on leveraging pair-wise relative Euclidean distance~\cite{xu-etal-2021-layoutlmv2,hwang-etal-2021-spatial}, we propose to incorporate pair-wise relative \textbf{hop distance}, which is defined as the shortest path between two entities on the \knn graph. Two entities could be in proximity in terms of Euclidean distance but not so in terms of hop distance. For example, in documents with complex layouts, it is common to have two entities that are close to each other in the Euclidean space, but there is a third entity positioned in between. This type of entity pair should be treated differently from pairs that are close to each other in both Euclidean and hop distances. 
In this case, the spatial attention mechanism based solely on the relative Euclidean distances between entity pairs is insufficient since it neglects this structural information.
We argue that the \knn graph structure is an effective way of capturing the structural information and propose to incorporate it as an inductive bias into the attention computation.

Intuitively, different hop distances should carry different weights in calculating pairwise attention.
We use $\phi_{(i,j)}$ to represent the hop distance between entity i and  j and $H$ to represent a learnable embedding lookup table based on the hop distance $\phi_{(i,j)}$. Inspired by DeBERTa~\cite{he2020deberta} and Transformer-XL~\cite{dai2019transformer}, we integrate the hop distance bias into attention as described in the following equations

\begin{dmath}
\label{eq_attn1}
    e_{ij} = [x_iW^Q(x_jW^K + H^Q_{\phi_{(i,j)}} + R^Q_{\sigma_{(i,j)}}) \\ + (H^K_{\phi_{(i,j)}} + R^K_{\sigma_{(i,j)}})x_iW^K] / \sqrt{d},
\end{dmath}

\begin{equation}
\label{eq_attn2}
    z_i = \sum_j a_{ij}(x_jW^V +  H^V_{\phi_{(i,j)}} + R^V_{\sigma_{(i,j)}}),
\end{equation}
where $\sigma_{(i,j)}$ is a concatenation of the relative Euclidean distance and angle between entity i and j, and $R$ is a learnable matrix. $H$ could be a learnable matrix or a lookup table that maps $\sigma_{(i,j)}$ to learnable embeddings. $e_{ij}$ is the attention weight between entity $i$ and $j$. $a_{ij}$ is calculated as the weight of $exp(e_{ij})$ in the exponential sum of all $e_{ik}$, as described in Eqn.\ref{eq_attn3}. 

\begin{equation}
\label{eq_attn3}
     a_{ij} = \frac{ exp(e_{ij})} {\sum_k exp(e_{ik})}.
\end{equation}

Similar to how pair-wise relative Euclidean distance is added to attention, we add pair-wise hop distance as three learnable weight matrices, two of which multiply with query and key vectors respectively while the remaining one is added to the value vector. 
 We further limit an entity's attention only to its local radius of neighborhood defined by the \knn graph. Specifically, we do not calculate $e_{ij}$ if the hop distance between entity $i$ and $j$ exceeds a certain threshold. This also strengthens the inductive bias as supported by our experiment results.

\subsubsection{Combinatorial Matching}
We hypothesize that combinatorial properties between field categories and entities can be leveraged as inductive biases to improve extraction performance.
Different from existing methods that treat the classification of each entity independently~\cite{ xu-etal-2021-layoutlmv2, hwang-etal-2021-spatial, lee-etal-2022-formnet}, we propose to treat the entity classification task as a set prediction problem to exploit the one-to-one mapping constraint, where one field has one and only one corresponding entity. 
The combinatorial assignment is described in Eqn.\ref{eq_hm}.

\begin{equation}
\label{eq_hm}
    \tau_{opt} = argmin_\tau \sum_i^N L_{match} (y_i^{label}, y_{\tau_{(i)}} ^ {pred}),
\end{equation}
where $\tau$ is an assignment, and $L_{match}$ is the matching cost. N is the number of entities in a document. In practice, N is often much larger than the number of entities of interest.
Therefore, we pad the number of ground truths to N in order to perform a one-to-one combinatorial assignment.
This can be done with the Hungarian algorithm in polynomial time \cite{kuhn1955hungarian, carion2020end,stewart2016end}.

\section{Datasets}
\label{sec:dataset}

Many real-world documents exhibit combinatorial properties, such as a one-to-one mapping between between its fields and entities.
However, existing public datasets do not cover documents with such properties~\cite{8892998, park2019cord, huang2019icdar2019}. To fill the gap, we release a new ID document dataset POI, and enhanced annotations of MIDV2020. We also verify our method on a private dataset PRV. All 3 datasets exhibit combinatorial properties.

In addition, we design the POI dataset to be template-rich with diverse languages. We also design the enhanced MIDV2020 with a difficult split such that templates in testing are unseen during training. 
BERT alone without spatial information can achieve above 90\% F1 on some existing datasets~\cite{hong2022bros, park2019cord, huang2019icdar2019}, indicating relative sufficiency of leveraging text information alone. Yet in many real-world use cases, using text alone is insufficient.
This motivates us to work on more challenging datasets where the exploitation of spatial information is important.
Dataset statistics are summarized in \ref{tab:doc_number_train_test} and Tab.~\ref{tab:dataset_stas}. More details are as follows.

\begin{table}[htbp]

\small
  \centering

    \begin{tabular}{lccc}
    \toprule
          & \multicolumn{1}{l}{\textbf{\#Train Doc.}} & \multicolumn{1}{l}{\textbf{\#Test Doc.}} \\
    \midrule
    POI    & 421   & 109 \\
     MIDV2020     & 500   & 200 \\
    PRV    & 3480  & 807 \\
    \bottomrule
    \end{tabular}%
    \caption{\label{tab:doc_number_train_test}Number of documents in training and testing.}
  
\end{table}%

\begin{table}[htbp]

\small
  \centering

    \begin{tabular}{lccc}
    \toprule
    \multicolumn{1}{p{3em}}{\textbf{Dataset}} & \multicolumn{1}{p{5.5em}}{\textbf{Avg \# of Ent. per Doc.}} & \multicolumn{1}{p{4.5em}}{\textbf{Total \# of Ent.}} & \multicolumn{1}{p{4.5em}}{\textbf{Total \# of Doc.}} \\
    \midrule
    POI   & 31.79 & 16850 & 530 \\
    MIDV2020  & 32.85 & 23000 & 700 \\
    PRV   & 24.31 & 104245 & 4287 \\
    \bottomrule
    \end{tabular}%
  \caption{\label{tab:dataset_stas}Statistics of entity distribution in documents. Ent. stands for entities and Doc. stands for documents. }
\end{table}%

\paragraph{POI}
We collect and annotate 530 Proof-of-Identity documents from online sources. We will release this POI dataset which consists of 10 document types, 265 distinct templates, and 131 countries of origin. The template and language diversity of POI create a challenging task for document understanding. All images are specimens with dummy values. The document type distribution is shown in Tab.\ref{tab:doc_type_distribution}. 

There are 8 field categories in total: last name, first name, date of birth, date of issue, date of expiry, ID number, key, and others. Key represents entities that indicate the field names for the important entities (e.g. Last Name) that we are interested to extract. The first 6 field categories appear in each document image once and only once, creating a special mapping constraint unseen in other datasets. The last 2 field categories (key and others) are not subject to the constraint. 
In real-world applications, it is common to extract a set of entities from documents that have combinatorial properties between its field and entities. ID document information extraction is one such use case, where we only expect to extract one entity for each field category of interest. This one-to-one correspondence can be leveraged to improve classification performance.
Despite being a common task setting, we notice the lack of method exploration and innovation in this direction, due to the unavailability of such property among existing popular document datasets. More details about the dataset can be found in the Appendix.

\begin{table}[htbp]

\small
  \centering
    \begin{tabular}{lc}
    \toprule
    \textbf{Document Type}& \multicolumn{1}{l}{\textbf{\# Document}}\\
    \midrule
    Passport & 238 \\
    Driving License & 119 \\
    Travel Document & 109 \\
    ID    & 30 \\
    Resident Permit & 21 \\
    Seafarer ID & 10 \\
    Others & 3 \\
    \bottomrule
    \end{tabular}%
  \caption{\label{tab:doc_type_distribution}Distribution of document types in POI dataset}
\end{table}%

\paragraph{MIDV2020}
We utilize the 1000 synthesized ID documents from the initial MIDV2020 dataset~\cite{DBLP:journals/corr/abs-2107-00396} . These documents are generated from 10 templates, with 100 documents for each template. Each document image is annotated with a list of bounding box coordinates and field values. We find that only artificially generated entities, such as the values of names and ID numbers, are annotated, while entities that belong to the original templates, such as document title and field names are not. We proceed to annotate the remaining entities. The newly annotated ground truths of MIDV2020 will be released alongside POI. 
These enhanced annotations enable us to perform information extraction task in a setting that is closer to real-world application, where all texts recognized by the OCR engine are used. 
The train/test split we introduce for MIDV2020 is a split by countries, this ensures that the document templates in the training dataset are unseen in the testing dataset. 
The country split simulates real-world scenarios where the model extension to new countries or new versions of documents is needed. More details can be found in the Appendix.

\paragraph{PRV}
Since POI and MIDV2020 only contain specimens or artificially generated images, we run our model on a private dataset (named PRV) that mostly consists of US driver licenses. 
The documents are protected by strict privacy requirements and massive human annotations are not available as raw images are inaccessible. Therefore, we build automatic fuzzy labeling to annotate the ground truth.

\paragraph{Comparison on Datasets}
POI exhibits better template and language diversity. POI contains 265 templates from 131 countries, while MIDV2020 has 10 templates from 10 countries. The number of templates in PRV is unknown due to privacy-related limitations. In addition, POI consists of templates in a multitude of languages, whereas MIDV2020 and PRV dataset lack such diversity. Texts in POI and MIDV2020 are made up largely by artificial text which is more readable and clearer, while PRV contains real texts. POI and PRV samples are split randomly. Since MIDV2020 has only 10 templates, we split the samples by country to make the task more challenging. PRV is the easiest dataset among the three due to its lingual monotony and random split.

\begin{table*}[htbp]
  \centering
  \resizebox{\linewidth}{!}{
    \begin{tabular}{crccccccccc}
    \toprule
    \multirow{2}[4]{*}{\textbf{Dataset}} & \multicolumn{1}{c}{\multirow{2}[4]{*}{\textbf{Method}}} & \multicolumn{6}{c}{\textbf{F1 Score}}         & \multirow{2}[4]{*}{\textbf{Input Modality}} & \multicolumn{2}{c}{\textbf{\#Parameters}} \\
\cmidrule{3-8}\cmidrule{10-11}          &       & \textbf{L.Name} & \textbf{F.Name} & \textbf{DoB} & \textbf{DoI} & \textbf{DoE} & \textbf{ID No.} &       & \textbf{Trainable} & \multicolumn{1}{c}{\textbf{Total}} \\
    \midrule
    POI   & BERT\textsubscript{BASE} & 67.90 & 72.73 & 92.11 & 70.78 & 69.06 & 78.70 & text  & 110 M & 110 M \\
          & GCN   & 45.35 & 56.08 & 85.62 & 62.37 & 62.32 & 70.65 & text + layout & 31.5 K & 22.7M \\
          & LayoutLM\textsubscript{BASE} & 87.03 & 86.88 & 93.93 & 86.23 & 87.72 & 83.12 & text + layout & 110 M & 110 M \\
          & LayoutLMv2\textsubscript{BASE} & \textbf{90.58} & \textbf{89.26} & 96.00 & 94.22 & 92.59 & 88.16 & text + layout + image & 199 M & 199 M \\
          & SPADE & 73.73 & 78.63 & 90.09 & 89.59 & 90.27 & 83.98 & text + layout & 128 M & 128 M \\
          & BROS\textsubscript{BASE} & 82.39 & 82.76 & 94.16 & 91.41 & 88.32 & 83.18 & text + layout & 109 M & 109 M \\
\cmidrule{2-11}          & KNN-former & 83.57 & 82.18 & \textbf{98.37} & \textbf{95.89} & \textbf{94.48} & \textbf{90.06} & text + layout & 0.5 M & 23.2M \\
    \midrule
    MIDV2020 & BERT\textsubscript{BASE} & 40.61 & 52.89 & \textbf{100.00} & 85.29 & 80.00 & 55.62 & text  & 110 M & 110 M  \\
          & GCN   & 32.03 & 43.09 & 99.50 & 99.00 & 79.76 & 43.82 & text + layout & 31.5 K & 22.7M \\
          & StructuralLM\textsubscript{LARGE} & 25.13 & 11.83 & \textbf{100.00} & 89.29 & 91.53 & \textbf{99.50} & text + layout & 355 M  & 355 M \\
          & LayoutLM\textsubscript{BASE} & 47.65 & 15.10 & \textbf{100.00} & 97.96 & 80.16 & 67.97 & text + layout & 110 M & 110 M \\
          & LayoutLMv2\textsubscript{BASE} & 47.54 & 49.91 & 87.15 & 97.56 & 77.24 & 94.18 & text + layout + image & 199 M & 199 M \\
          & SPADE & 48.91 & 45.54 & 79.90 & 63.47 & 60.85 & 60.34 & text + layout & 128 M & 128 M \\
          & BROS\textsubscript{BASE} & 23.31 & 23.78 & 98.50 & 70.83 & 18.27 & 85.39 & text + layout & 109 M & 109 M \\
\cmidrule{2-11}          & KNN-former & \textbf{87.88} & \textbf{54.26} & \textbf{100.00} & \textbf{100.00} & \textbf{95.21} & 69.65 & text + layout & 0.5 M & 23.2M \\
\midrule
    PRV   & BERT\textsubscript{BASE} & 71.32 & 76.39 & 97.72 & 88.78 & 86.22 & 87.21 & text  & 110 M  & 110 M \\
          & GCN   & 66.32 & 81.97 & 97.59 & 89.53 & 87.90 & 89.38 & text + layout & 31.5 K & 22.7M \\
          & StructuralLM\textsubscript{LARGE} & 93.72 & 93.27 & \textbf{99.56} & 98.86 & 99.21 & 97.86 & text + layout & 355 M & 355 M \\
          & LayoutLM\textsubscript{BASE} & \textbf{95.36} & 94.71 & 99.17 & 98.76 & 98.61 & 97.85 & text + layout & 110 M & 110 M \\
          & LayoutLMv2\textsubscript{BASE} & 95.26 & 95.31 & 99.52 & 99.29 & 99.36 & \textbf{98.82} & text + layout + image & 199 M & 199 M \\
          & SPADE & 65.61 & 70.65 & 98.70 & 98.10 & 96.43 & 92.48 & text + layout & 128 M & 128 M \\
          & BROS\textsubscript{BASE} & 93.52 & 91.68 & 99.00 & 98.44 & 97.53 & 97.91 & text + layout & 109 M & 109 M \\
\cmidrule{2-11}          & KNN-former & 92.03 & \textbf{96.81} & 91.22 & \textbf{99.68} & \textbf{99.47} & 98.76 & text + layout & 0.5 M & 23.2M \\
    \bottomrule
    \end{tabular}%
}
  \caption{\label{tab:main_table}Entity-level F1 score of \kf compared to baselines. Column L.Name, F.Name, DoB, DoI, DoE and ID No. correspond to results of Last Name, First Name, Date of Birth, Date of Issue, Date of Expiry, and ID Numbers. GCN and \kf have additional 22.7 M fixed parameters since we employed a light-weighted 6-layer sentence transformer~\cite{reimers-2019-sentence-bert} to get the text embeddings.} 
\end{table*}%

\section{Experiments}

In this section, we conduct extensive experiments to evaluate our proposed \kf on aforementioned datasets. We first compare our results with several baselines in Sec.~\ref{sec:main_results}. Then in Sec.~\ref{sec:eva_gen_abi}, we evaluate the generalization ability of our method on unseen templates. We then conduct ablation studies in Sec.\ref{sec:abl_study} and Sec.\ref{sec:k_ablation} to assess the effects of each component in \kf and the impact of different \textit{K} in the \knn graph.

\subsection{Comparison with Baselines on Multiple Datasets}
\label{sec:main_results}

We first evaluate the performance of \kf against multiple competitive methods. We choose base models for most of the baselines, because these are closest to \kf in terms of the number of parameters. Brief description of baseline models as well as the implementation details of all the models can be found in Sec.~\ref{sec:implement}. We do not have results for StruturalLM on POI dataset because of an OOV error.

Tab.\ref{tab:main_table} shows the entity-level classification performance. 
The results show that our method outperforms the baselines on most entity types across various datasets. In particular, \kf outperforms LayoutLMv2\textsubscript{BASE}, a state-of-the-art model that uses additional image features. We also observe that BERT performs poorly on these datasets, indicating the importance of exploiting spatial information.

Secondly, as shown in Trainable Param column in Tab.\ref{tab:main_table}, \kf is highly parameter-efficient. All baselines except GCN have more than 100 million trainable parameters, while \kf has only $0.5$ million and is magnitudes smaller than competing methods. Even after adding the sentence transformer, \kf has only 23.2 million parameters, still 5x smaller than baselines.
The parameter efficiency has 4 benefits. First, it contributes to learning and inference time efficiency, with details illustrated in \ref{sec:runtime}.
Second, it allows for faster fine-tuning on new datasets and domains, especially in real-world use cases when training datasets are big and re-training requirements are frequent. Third, smaller model size and faster inference time make mobile deployment more feasible. Fourth, training, fine-tuning and inferring smaller models reduces power consumption and carbon footprint. Despite the smaller model size, \kf achieves comparable or better performance across datasets.

Thirdly, we observe that \kf underperforms both LayoutLM\textsubscript{BASE} and LayoutLMv2\textsubscript{BASE} for name related entities in both POI and PRV datasets. The robustness of the two baselines in predicting names could be attributed to their extensive pre-training. The two baselines learn common names in pre-training, enabling them to predict names correctly regardless of context. However, despite no extensive pre-training, \kf still outperforms BROS and StructuralLM which are also pre-trained on 11 million documents.

Fourthly, we observe all methods suffer performance degradation on MIDV2020, compared to the other two datasets. This is because in MIDV2020, training and testing documents are split by countries, templates in testing are not seen during training. In addition, MIDV2020 has only 6 templates in training data, which easily leads to overfitting. Detailed discussion on the generalization ability can be found in Sec.~\ref{sec:eva_gen_abi}. 
We find that BERT outperforms several baselines with spatial modelling on names, this may be due to overfitting to limited number of training templates. We notice that our method do not perform well on id number entity. We conducted manual inspection on several error cases, and find that in many documents there exist two different types of id numbers(see Fig.~\ref{fig:example_doc}(b)), but only one of them is labeled as id number according to the provided annotations. Our model sometimes predicts the other one as id number. This also explains the poor performance on id number for some other baselines.

Lastly, we notice that on the PRV dataset, \kf performs poorly on DoB field, underperforming even GCN. \kf's performance on DoB drops after combinatorial matching, despite an overall increase macro average F1.
This could be due to the presence of noise in groundtruth, since this dataset is annotated by automatic fuzzy labeling logic. 
Manual examination of a few documents confirms our hypothesis.

\subsection{Evaluation of generalization ability on unseen templates}
\label{sec:eva_gen_abi}

\begin{figure}[htbp]
\centering
\includegraphics[width=1\linewidth]{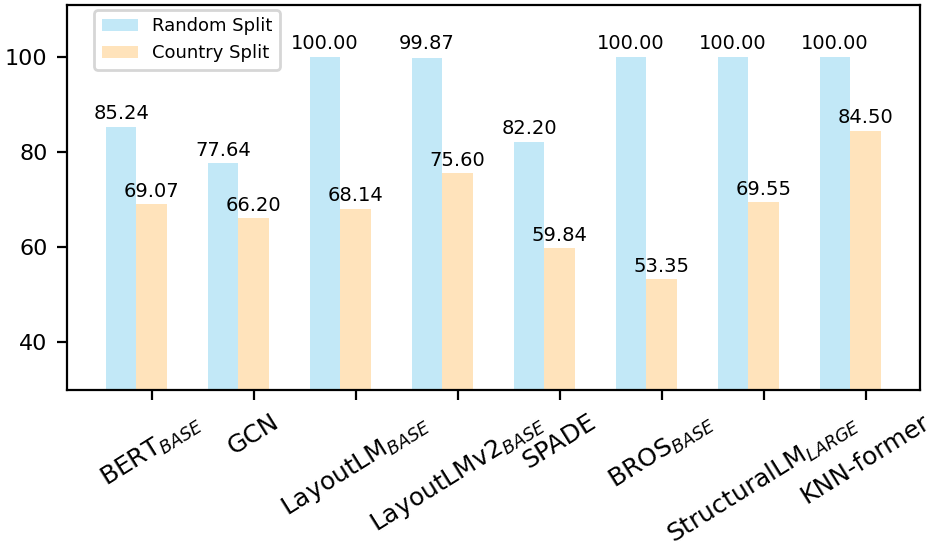}
\caption{\label{fig:random_country_comparisons_midv}Macro average F1 scores of \kf and various baseline models under random split and country split on MIDV2020 dataset.}
\end{figure}

To assess the generalization capability of our model, we test and compare our model with other competitive baselines on MIDV2020 dataset using two train/test settings: random split and split by country . The country split is a more difficult setting as the templates in testing are unseen during training. Intuitively, we would expect a decline in performance as compared to the random split setting.  Fig.~\ref{fig:random_country_comparisons_midv} shows the Macro average F1 scores comparison of \kf and multiple baselines under both the random split and the country split. 

We observe across-the-board performance degradation for all methods after switching from random split to country split. However, the drop is least significant on \kf, enabling it to achieve 10\% higher F1 than the best baseline.

These experiments indicate that our method is more robust and generalizes better to unseen templates as compared to existing baseline models. This is helpful in real-world applications where models frequently encounter new types of documents.

\subsection{Effects of each component in \kf}
\label{sec:abl_study}

\begin{table}[htbp]

  \resizebox{1.0\linewidth}{!}{
      \begin{tabular}{ll}
        \toprule
        \textbf{Model} &\textbf{F1}\\
        \midrule
        \text{\kf}&90.76\\ 
        \text{\quad(-)\knn hop attention}&88.33 (-2.43)\\ 
        \shortstack{\quad(-)Local attention based on \knn hop \\ \& (-)\knn hop attention}&85.67 (-5.09)\\
        \text{\quad(-)Relative Euclidean distance \& angle attention}&87.17 (-3.59)\\
        \shortstack{\quad(-)Relative Euclidean distance \& angle attention \\ \& (-)\knn hop attention}&86.67 (-4.09)\\
        \text{\quad(-)Combinatorial Matching}&88.16 (-2.60)\\
        \text{\quad(+)Absolute positional encoding}&86.33 (-4.43)\\
      \bottomrule
      \end{tabular}
  }
  \caption{\label{tab:alation} Ablation results on POI dataset. (-) indicates the component is absent compared to \kf, (+) indicates the component is additional.}
\end{table}

To better understand how \kf works, we ablatively study the effects of each component and report the results in Tab.~\ref{tab:alation}. Entity-level detailed results can be found in the Appendix.

Firstly, we observe a 2.43\% drop in performance with the removal of \knn hop attention and an even bigger 5.09\% drop when local attention is removed together with \knn hop attention. This demonstrates that the \knn graph-based inductive bias is effective in capturing the structural information between document entities. It also shows that local attention, the practice of masking out attention weights when the hop distance between two entities exceeds a pre-defined threshold, further strengthens the inductive bias. 

Secondly, we observe that the commonly used spatial inductive bias based on the pairwise relative Euclidean distance and angle also plays an important role. When both relative Euclidean distance attention and \knn hop attention are absent, there is a 4.09\% drop in performance, an additional decrease of 1.66\% compared to when only \knn hop attention is ablated(2.43\%).
The overlap of performance drop suggests some information are captured by both Euclidean distance and hop distance, as some pairs are similarly close/far from each other as measured in both distances. However, each distance also complements the other by capturing additional information. For example, two pairs could carry different importance despite having the same Euclidean distance, due to the presence or absence of other entities positioned between the pairs, signifying the importance of hop distance.

Thirdly, we notice that the F1 score drops drastically by 4.76\% when combinatorial matching is ablated. This demonstrates the important contribution of combinatorial matching, as the datasets we experiment on are all subject to a special one-to-one mapping constraint between fields and entities. Combinatorial matching enables our method to treat entity classification as a set prediction problem, instead of predicting each entity's class independently, which enhances our model robustness.

Lastly, we observe that there is a 4.43\% drop in performance when absolute positional encoding is added. Previous works~\cite{hwang-etal-2021-spatial} have similar findings that adding absolute positional encoding is not helpful, especially when the test set contains a diverse set of unseen templates. In our experiments, adding absolute positional encoding improves performance in training but generalizes poorly in testing. 

\subsection{Impact of different \textit{K} in the \knn graph}
\label{sec:k_ablation}

\begin{table}[htbp]

    \small
    \centering
    \begin{tabular}{rrrr}
    \toprule
    \multicolumn{1}{l}{\textbf{\#K}} & \multicolumn{1}{l}{\textbf{(+) H (-) R}} & \multicolumn{1}{l}{\textbf{(-) H (+) R}} & \multicolumn{1}{l}{\textbf{(+) H (+) R}} \\
    \midrule
    2     & 90.67 & 89.33 & 89.50 \\
    5     & 88.74 & 90.23 & 89.51 \\
    \bottomrule
    \end{tabular}%
      \caption{\label{tab:k_impact}Impact of number of K in \kf on POI dataset. (+) indicates presence, (-) indicates absence. H refers to the KNN hop attention. R refers to relative Euclidean distance and angle attention.}

\end{table}%
To further study the effect of how the hyper-parameter of the \knn graph affects the performance, we conduct experiments with different values of \textit{K} on the POI dataset. As shown in Tab.~\ref{tab:k_impact}, the 2-NN graph achieves the best performance when \knn-based hop distance is used and relative Euclidean distance is removed. This is because when only 2 nearest entities are counted as an entity's first-hop neighbors, the correlation between hop distance and entity pair's importance is pronounced. However, a 5-NN graph achieves the best performance when \knn-based hop distance is ablated and only relative Euclidean distance is used. This is because the information of who is an entity's 5 nearest neighbors is less useful in documents with an average of 31.79 annotated bounding boxes per file. Models with 2-NN and 5-NN graphs underperform the 4-NN graph in the POI dataset, underscoring the importance of choosing the correct \knn graph hyper-parameter for different datasets.

\subsection{Runtime Comparison}
\label{sec:runtime}
In addition to performance evaluation, we also evaluate the runtime of our model against competitive baselines. For fair comparison, we report the total runtime of sentence transformer plus \kf, since \kf uses sentence transformer for text embeddings. In fact, the sentence transformer takes up half of the time in our pipeline.

\begin{table}[htbp]

\small
  \centering
    \begin{tabular}{lrr}
    \toprule
    \textbf{Model}& \textbf{Single} &\textbf{Batch}\\
    \midrule
           LayoutLM\textsubscript{BASE} &19.61 &237.90\\
           LayoutLMv2\textsubscript{BASE}  &56.64 &2941.32\\
           SPADE  & 39.47 & 6091.52 \\
           BROS\textsubscript{BASE} &23.45&646.65\\
          \midrule
         \kf  &22.60&77.57\\
    \bottomrule
    \end{tabular}%
  \caption{\label{tab:runtime}Runtime comparison with baselines. Time taken is reported in milliseconds.}
\end{table}%

We first measure the runtime to process a single document for each method. As shown in Tab.~\ref{tab:runtime}, time taken for sentence encoder plus KNN-former is comparable to LayoutLM  and BROS, and is faster than SPADE, LayoutLMv2. We run StruturalLM(written in tensorflow1.14) on CPU due to cuda version mismatch, hence there is no speed measurement.

Moreover, our method allows for significantly larger batch sizes because of the smaller model size. Therefore, runtime for documents in batch is significantly faster than the baselines. Running with maximum possible batch size for each model using a 16GB V100 GPU, \kf is significantly faster than the rest, as shown in Tab.~\ref{tab:runtime}.
This experiment demonstrates that our model is advantageous when faster execution time is desirable, and this could be attributed to the lightweight property of our model.

\section{Conclusion}
We propose \kf, a parameter-efficient transformer-based model for document entity classification. \kf uses \knn Hop Attention, a new attention mechanism that leverages \knn graph-based inductive bias to capture structural information between document entities. \kf utilizes combinatorial matching to perform set prediction. We also release POI, a template-rich ID document dataset subject to combinatorial constraints. 
Experiments show that \kf outperforms baselines in entity classification across various datasets.

\section*{Limitations}

We identify the following limitations in this work. 
First, the robust performance of baseline methods that leverage image features~\cite{appalaraju2021docformer} testifies to the importance of visual cues. The inclusion of image features to \kf might contribute to better performance. 
Second, unlike models that perform extensive pre-training \cite{10.1145/3394486.3403172,xu-etal-2021-layoutlmv2}, \kf might lack generic domain knowledge. 
Third, \kf uses a vanilla sentence transformer to get the text embedding inputs. The sentence transformer model is pre-trained and not fine-tuned on the new datasets. An end-to-end training pipeline that jointly trains the text encoding model and \kf could lead to better results.
Fourth, there are many design choices we did not explore, such as applying attention directly at the token level and pooling representations at the end. Lastly, KNN-Former, along with all baselines used in this work, are subject to OCR failure. All models consume OCR outputs such as bounding box coordinates and texts. In the case of OCR failure, where one bounding box is detected as two or two boxes are merged as one, models that consume OCR results are less likely to make correct predictions.

\section*{Ethics Statement}
This work has obtained clearance from author's institutional review board. The annotators for POI and MIDV2020 are all paid full-time interns and researchers hired by our institute, whose compensation are determined based on the the salary guidelines of our institute. Among the datasets and annotations released, POI only contains specimens with dummy values, while MIDV is a synthetic dataset. External data are accessed and used in compliance with fair use clauses. We conduct experiments on the private dataset PRV in a secure data zone with strict access control, using auto-labeling scripts for annotations.  

\section*{Acknowledgement}
This research is partly supported by the SRG grant id: T1SRIS19149 and the Ministry of Education, Singapore, under its AcRF Tier-2 grant (Project no. T2MOE2008, and Grantor reference no. MOET2EP20220-0017). Any opinions, findings, conclusions, or recommendations expressed in this material are those of the author(s) and do not reflect the views of the Ministry of Education, Singapore.

\bibliography{anthology,custom}
\bibliographystyle{acl_natbib}

\appendix

\section{Appendix}

\subsection{Implementation details}
\label{sec:implement}
We briefly describe the baseline models as well as detailed implemetation details of all models in this section.
\begin{itemize} 
    \item \textbf{BERT\textsubscript{BASE}}~\cite{devlin-etal-2019-bert}: We use the pre-trained BERT base model for token classification. 
    \item \textbf{GCN}~\cite{kipf2016semi}: We use sentence transformer~\cite{reimers-2019-sentence-bert} to get the embeddings of text inputs and use them as the node features for the constructed \knn graph. Then we train a 2-layer graph convolutional network to classify the nodes/entities.
    \item \textbf{LayoutLM\textsubscript{BASE}}~\cite{10.1145/3394486.3403172}: LayoutLM is a transformer-based model for document image understanding. It is pre-trained on IIT-CDIP Test Collection with 11
    million scanned images. 
    \item \textbf{LayoutLMv2\textsubscript{BASE}}~\cite{xu-etal-2021-layoutlmv2}: In addition to LayoutLM, the LayoutLMv2 adds a new multi-modal task during pre-training to take in the visual cues and incorporates a novel spatial-aware self-attention mechanism. 
    \item \textbf{StructuralLM\textsubscript{LARGE}}~\cite{li2021structurallm}: On top of LayoutLM, Structural LM uses cell position for each word, and introduces a new pre-training task that predicts the cell position. It is also pre-trained on the IIT-CDIP dataset.
    \item \textbf{SPADE} \cite{hwang-etal-2021-spatial}: SPADE builds a directed graph of document entities and extracts and parses the spatial dependency using both linguistic and spatial information. 
    \item \textbf{BROS}~\cite{hong2022bros}: Similar to LayoutLM, BROS is also pre-trained on the IIT-CDIP dataset, but with a different area masking pre-training task, and a different method to encode the 2D positions of bounding boxes.
    \item \textbf{DocFormer}~\cite{appalaraju2021docformer}: DocFormer is a multi-modal transformer that takes in both text and visual cues. It proposes a multi-modal attention mechanism and is pre-trained with several tasks involving both text and image input.
\end{itemize}

All models are trained on 16G V100 GPUs and implemented with Pytorch, except for StructuralLM\textsubscript{LARGE}, for which we use their official repository~\footnote{https://github.com/alibaba/AliceMind/StructuralLM} that is implemented in Tensorflow1.14 and we train it on cpu because of cuda version mismatch. 
We use APIs open-sourced by Huggingface~\footnote{https://huggingface.co} for Bert, LayoutLM\textsubscript{BASE} and LayoutLMv2\textsubscript{BASE}.
SPADE is implemented using the official implementation released by ClovaAI\footnote{https://github.com/clovaai/spade}. BROS is implemented using their released official repository~\footnote{https://github.com/clovaai/bros}. Only text inputs are passed to BERT\textsubscript{BASE} for classification while bounding box coordinates are neglected. Results are obtained after training for 100 epochs. We trained the SPADE model for 10 to 20 hours up to 1000 epochs depending on the datasets. All settings of  LayoutLM\textsubscript{BASE} and LayoutLMv2\textsubscript{BASE} are from the authors. For BROS, we use the same tokenizer as LayoutLM, same learning rate in their paper and fine-tuned BROS on each dataset for at least 100 epochs, and made sure it converged. We report results for epoch 80. For StructuralLM\textsubscript{LARGE}, we were only partially successful to reproduce it due to OOV error when running on POI dataset. In addition, this is the only baseline that we use the large version because there was an error with the base version. we train the model with 25 epochs with all other hyperparameters following their paper.
We reproduced DocFormer from an unofficial repository~\footnote{https://github.com/shabie/docformer} since there is no official repository available. There is no released pretraining weights for DocFormer, but DocFormer uses plain ResNet50~\cite{he2016deep} as the first step for image feature extraction, and the language embedding weights are initialized with LayoutLMv1\textsubscript{BASE} pre-trained weights. We trained DocFormer for at least 100 epochs and used hyperparameters for fine-tuning setting mentioned in the paper. We report results for epoch 100.

For \kf, we use 8 layers, 8 heads, and 80 hidden dimensions for the architecture. Results are obtained after training for 400 epochs. We use a 6-layer sentence transformer to extract text features in for both \kf and GCN baseline model implementation. We use Adam optimizer with learning rate of 5e-3. We perform a grid search in choosing hyper-parameters, with learning rate in [5e-3, 1e-3, 5e-4], the number of layers in [4, 8], local attention threshold in [1,2,3], and the number of attention heads in [4,8]. 
To incorporate relative Euclidean distance and angle, we tried both real and quantized angles in our initial exploration and did not find a significant difference. We use real angle values throughout the experiments.
In the implementation of combinatorial matching, we choose class probabilities as matching cost following~\citep{carion2020end}. Despite no theoretical justification, they observe better performance than log probabilities. We conduct experiments comparing class and log probabilities but do not observe significant differences in POI dataset(<0.005\%). Reported results are the average performance of 3 runs. The sentence transformer we used is paraphrase-MiniLM-L6-v2 from hugging face. 

\begin{table*}[t]
  \centering
  \resizebox{0.75\linewidth}{!}{
    \begin{tabular}{clccccccc}
    \toprule
    \multirow{2}[4]{*}{\textbf{Dataset}} & \multicolumn{1}{c}{\multirow{2}[4]{*}{\textbf{Method}}} & \multicolumn{6}{c}{\textbf{F1 Score}}         & \multicolumn{1}{c}{\multirow{2}[4]{*}{\textbf{Trainable Param}}} \\
\cmidrule{3-8}          &       & \textbf{L.Name} & \textbf{F.Name} & \textbf{DoB} & \textbf{DoI} & \textbf{DoE} & \textbf{ID No.} &  \\
    \midrule
    MIDV  & BERT\textsubscript{BASE} & 72.09 & 81.35 & 100.00 & 92.99 & 88.48 & 76.52 & 110 M \\
          & GCN   & 51.48 & 61.66 & 98.68 & 91.59 & 88.55 & 73.90 & 31.5 K \\
          & StructuralLM\textsubscript{LARGE}&\textbf{100.00} & \textbf{100.00} & \textbf{100.00} & \textbf{100.00} & \textbf{100.00} & \textbf{100.00}  &355M\\
          & LayoutLM\textsubscript{BASE} & \textbf{100.00} & \textbf{100.00} & \textbf{100.00} & \textbf{100.00} & \textbf{100.00} & \textbf{100.00} & 110 M \\
          & LayoutLMv2\textsubscript{BASE} & 99.47 & 99.74 & \textbf{100.00} & \textbf{100.00} & \textbf{100.00} & \textbf{100.00} & 199 M \\
          & SPADE & 88.14 & 86.82 & 70.63 & 80.33 & 79.71 & 87.55 & 128 M \\
          & BROS\textsubscript{BASE} & \textbf{100.00} & \textbf{100.00} & \textbf{100.00} & \textbf{100.00} & \textbf{100.00} & \textbf{100.00} & 109 M \\
\cmidrule{2-9}          & \kf & \textbf{100.00} & \textbf{100.00} & \textbf{100.00} & \textbf{100.00} & \textbf{100.00} & \textbf{100.00} & 0.5 M \\
    \bottomrule
      \end{tabular}%
    }
\caption{\label{tab:random_split_result}Experimental Results on MIDV2020 Random Split.} 
\end{table*}

\begin{table*}[tbhp]
  \centering
  \resizebox{1\linewidth}{!}{
  \begin{tabular}{crccccccccc}
    \toprule
    \multirow{2}[4]{*}{\textbf{Dataset}} & \multicolumn{1}{c}{\multirow{2}[4]{*}{\textbf{Method}}} & \multicolumn{6}{c}{\textbf{F1 Score}}         & \multirow{2}[4]{*}{\textbf{Input Modality}} & \multicolumn{2}{c}{\textbf{\#Parameters}} \\
\cmidrule{3-8}\cmidrule{10-11}          &       & \textbf{L.Name} & \textbf{F.Name} & \textbf{DoB} & \textbf{DoI} & \textbf{DoE} & \textbf{ID No.} &       & \textbf{Trainable} & \multicolumn{1}{c}{\textbf{Total}} \\
    \midrule
    POI   & \multicolumn{1}{r}{\multirow{4}[2]{*}{DocFormer\textsubscript{BASE}}} & 78.22 & 78.87 & 95.15 & 90.99 & 91.82 & 81.65 & \multirow{4}[2]{*}{text + layout + image} & \multirow{4}[2]{*}{110M} & \multirow{4}[2]{*}{110M} \\
    PRV   &       & 78.21 & 84.86 & 98.17 & 96.42 & 97.38 & 91.89 &       &       &  \\
    MIDV2020 (random split) &       & 100.00 & 100.00 & 100.00 & 100.00 & 100.00 & 100.00 &       &       &  \\
    MIDV2020 (country split) &       & 1.50  & 0.00  & 0.00  & 1.91  & 0.00  & 0.00  &       &       &  \\
    \bottomrule
    \end{tabular}%
    }
     \caption{\label{tab:df_res}Experimental Results on DocFormer.} 
\end{table*}

\subsection{Experimental Results on MIDV2020 random split}
Tab \ref{tab:random_split_result} shows the additional experimental results on MIDV2020 random split.  Column L.Name, F.Name, DoB, DoI, DoE and ID No. correspond to results of Last Name, First Name, Date of Birth, Date of Issue, Date of Expiry, and ID Numbers. GCN and \kf have additional 22.7 M fixed parameters since we employed a light-weighted 6-layer sentence transformer~\cite{reimers-2019-sentence-bert} to get the text embeddings. MIDV dataset has 10 templates, and each template has 100 images. As a result, this random split is an easy setting where performance results are generally good. BERT\textsubscript{BASE} still produces relatively poor performance, which reiterate the point that spatial information is important.

\subsection{Experimental Results on DocFormer}
Tab \ref{tab:df_res} shows the experimental results of DocFormer on various datasets. On POI, PRV dataset and MIDV2020 dataset random split, DocFormer performs reasonably well. On POI dataset, it only falls behind LayoutLMv2\textsubscript{BASE} and \kf; on PRV dataset, it outperforms BERT\textsubscript{BASE}, GCN and SPADE; on MIDV2020 dataset random split, it achieves 100\% F1 score for every field like \kf, StructuralLM\textsubscript{LARGE}, LayoutLM\textsubscript{BASE} and BROS\textsubscript{BASE}. However, on MIDV2020 dataset country split, we cannot get reasonable performance for DocFormer although we made sure our training was converged. 

We also measured the runtime of DocFormer, results shown in Tab.~\ref{tab:runtime_doc}.
\begin{table}[htbp]

\small
  \centering
    \begin{tabular}{lrr}
    \toprule
    \textbf{Model}& \textbf{Single} &\textbf{Batch}\\
    \midrule
           LayoutLM\textsubscript{BASE} &19.61 &237.90\\
           LayoutLMv2\textsubscript{BASE}  &56.64 &2941.32\\
           SPADE  & 39.47 & 6091.52 \\
           BROS\textsubscript{BASE} &23.45&646.65\\
           DocFormer\textsubscript{BASE} &71.57&7485.10\\
          \midrule
         \kf  &22.60&77.57\\
    \bottomrule
    \end{tabular}%
  \caption{\label{tab:runtime_doc}Runtime comparison with baselines. Time taken is reported in milliseconds.}
\end{table}%

\subsection{POI Dataset Details}

All images are publicly available specimen ID documents and do not contain information about real persons. Despite that, due to the sensitivity of the subject and increasing societal concerns about the role artificial intelligence should play in protecting people's privacy, we will only release the annotated JSON file instead of the actual images to comply with fair use of specimens. 

We store a list of objects in the annotated file; each object contains annotations for an image. The annotations include bounding box coordinates, text, and category. 

The released dataset is subject to fair use clause and should only be used for academic purposes. 

We implement quality control during the annotation process by having annotators cross-check each other's results to affirm the correctness of labels.

\begin{figure*}[htbp]
\begin{subfigure}{.33\textwidth}
  \centering
  \includegraphics[width=0.91\linewidth]{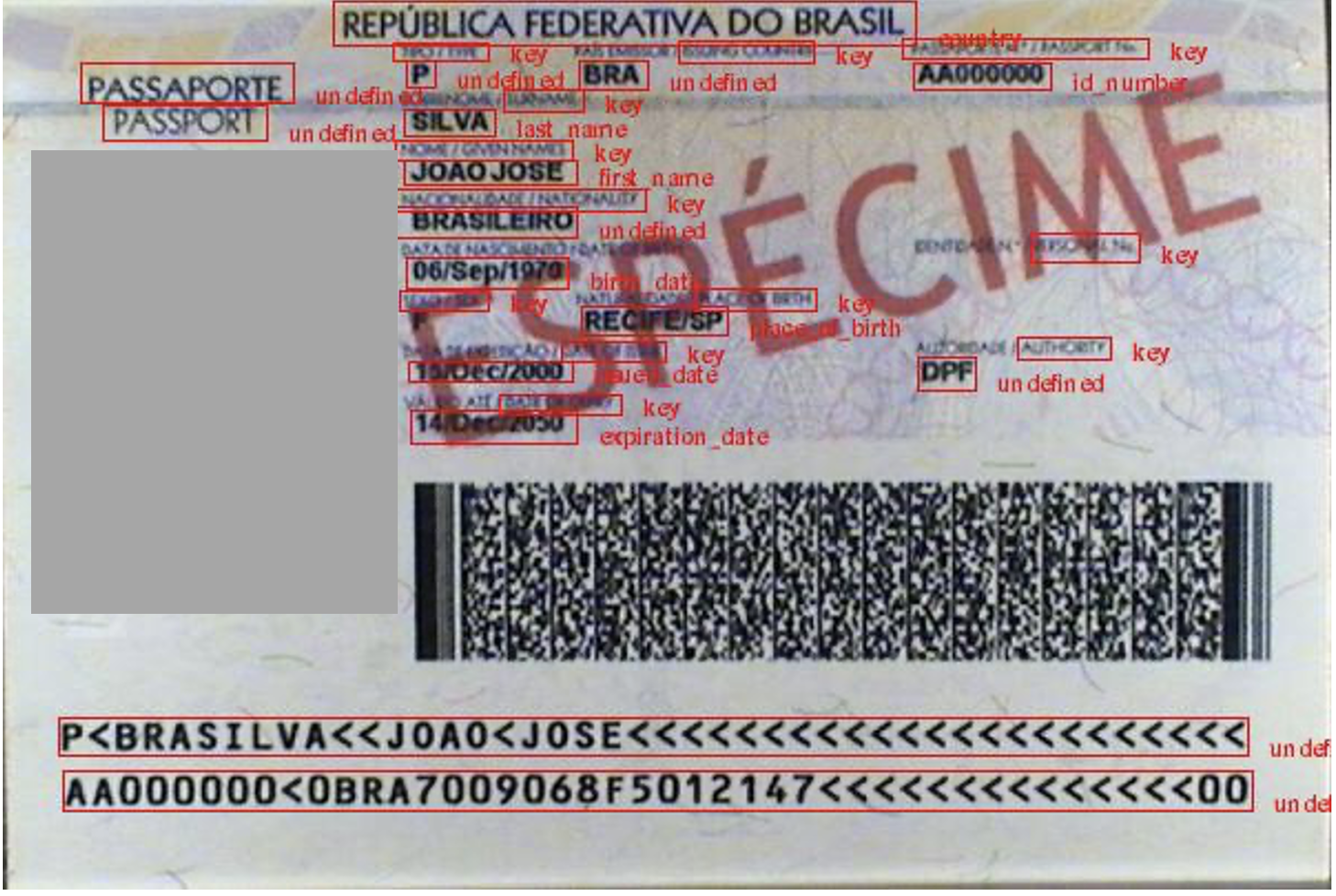}
  \caption{POI document}
  \label{fig:sfig1}
\end{subfigure}%
\begin{subfigure}{.33\textwidth}
  \centering
  \includegraphics[width=0.9\linewidth]{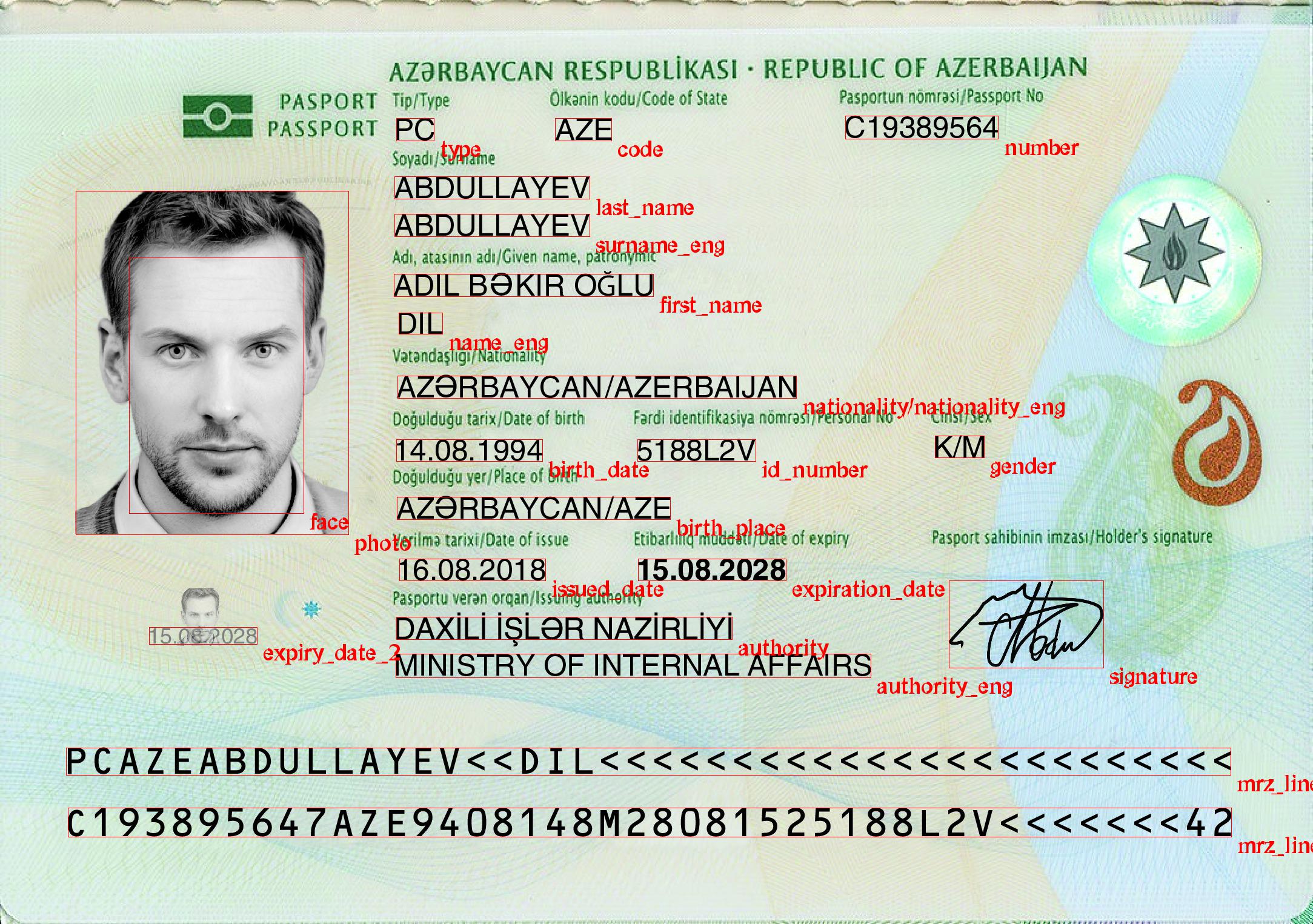}
  \caption{Original MIDV2020 document}
  \label{fig:sfig2}
\end{subfigure}
\begin{subfigure}{.33\textwidth}
  \centering
  \includegraphics[width=0.9\linewidth]{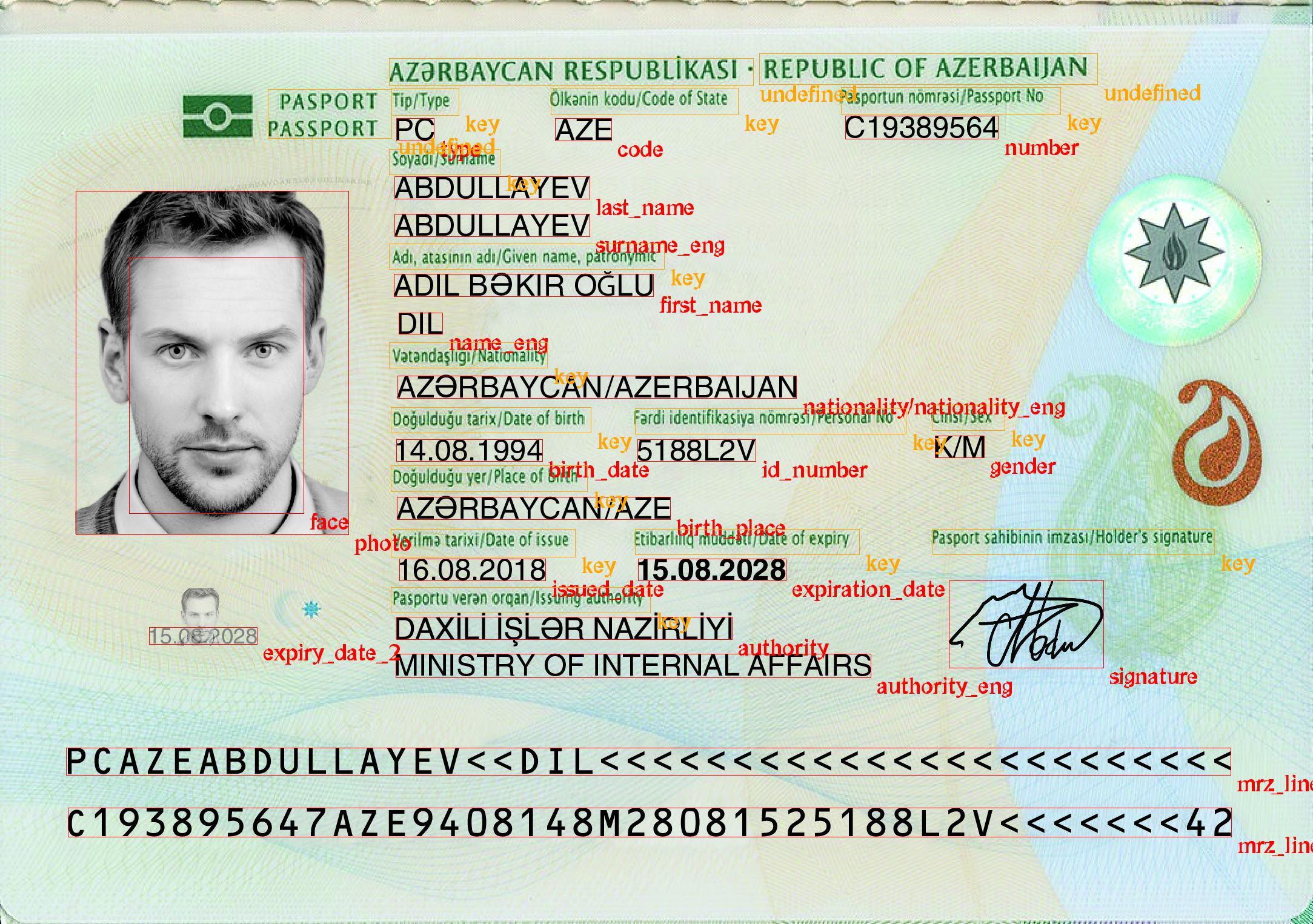}
  \caption{Enhanced MIDV2020 document}
  \label{fig:sfig3}
\end{subfigure}

\caption{Example documents with bounding boxes and annotations. There is only one entity box corresponding to one field of interest.}
\label{fig:example_doc}
\end{figure*}
\subsection{Sample documents of POI and MIDV2020}
In Fig.~\ref{fig:example_doc}, we show samples documents with bounding boxes and annotations.

\subsection{PRV Dataset Details} 
Since POI and MIDV2020 only contain specimens or artificially generated images, we run our model on a private (PRV) dataset that consists of actual ID documents. The documents are protected by strict privacy requirements and massive human annotations are not available as raw images are inaccessible. Therefore, we build automatic labeling to annotate the ground truth. Specifically, we map personal information in the existing database to OCR-ed text outputs. The matched bounding box is classified as the corresponding entity if a match is found. All bounding boxes that are not matched are classified as `others'.

\end{document}